\documentclass[journal]{IEEEtran}
\usepackage{cite}
\usepackage{amsmath,amssymb,amsfonts}
\usepackage{algorithmic}
\usepackage{algorithm}
\usepackage{graphicx}
\usepackage{textcomp}
\usepackage{amsmath}
\usepackage{mathtools}
\usepackage{bm}
\usepackage{url}
\usepackage{subcaption}
\usepackage{booktabs}
\usepackage{multirow}    
\usepackage{tabularx}
\usepackage{pifont}
\usepackage{hyperref}
\usepackage{xcolor}

\begin{document}

\title{CDPDNet: Integrating Text Guidance with Hybrid Vision Encoders for Medical Image Segmentation}

\author{Jiong Wu, Yang Xing, Boxiao Yu, Wei Shao, and Kuang Gong$^*$
\thanks{Jiong Wu, Xing Yang,  Boxiao Yu, and Kuang Gong are with the J. Crayton Pruitt Family Department of Biomedical Engineering, University of Florida, Gainesville, FL, 32611, USA. (e-mail: wujiong@ufl.edu; e-mail: yxing2@ufl.edu; e-mail: boxiao.yu@ufl.edu; email: KGong@bme.ufl.edu).}
\thanks{Wei Shao is with the Department of Medicine, University of Florida, Gainesville, FL, 32611, USA. (e-mail: weishao@ufl.edu).}}

\markboth{Journal of \LaTeX\ Class Files,~Vol.~14, No.~8, August~2015}%
{Shell \MakeLowercase{\textit{et al.}}: Bare Demo of IEEEtran.cls for IEEE Journals}

\maketitle

\begin{abstract}
Most publicly available medical segmentation datasets are only partially labeled, with annotations provided for a subset of anatomical structures. When multiple datasets are combined for training, this incomplete annotation poses challenges, as it limits the model's ability to learn shared anatomical representations among datasets. Furthermore, vision-only frameworks often fail to capture complex anatomical relationships and task-specific distinctions, leading to reduced segmentation accuracy and poor generalizability to unseen datasets. In this study, we proposed a novel CLIP-DINO Prompt-Driven Segmentation Network (CDPDNet), which combined a self-supervised vision transformer with CLIP-based text embedding and introduced task-specific text prompts to tackle these challenges. Specifically, the framework was constructed upon a convolutional neural network (CNN) and incorporated DINOv2 to extract both fine-grained and global visual features, which were then fused using a multi-head cross-attention module to overcome the limited long-range modeling capability of CNNs. In addition, CLIP-derived text embeddings were projected into the visual space to help model complex relationships among organs and tumors. To further address the partial label challenge and enhance inter-task discriminative capability, a Text-based Task Prompt Generation (TTPG) module that generated task-specific prompts was designed to guide the segmentation. Extensive experiments on multiple medical imaging datasets demonstrated that CDPDNet consistently outperformed existing state-of-the-art segmentation methods. Code and pretrained model are available at: \url{https://github.com/wujiong-hub/CDPDNet.git}.
\end{abstract}

    
\IEEEpeerreviewmaketitle

\section{Introduction}
\label{sec:introduction}
\IEEEPARstart{M}{edical} image segmentation plays a crucial role in disease diagnosis, treatment planning, and biomedical research~\cite{sahiner2019deep,park2021computer,kashyap2025automated}. However, the labor-intensive process of delineating organs and tumors across large datasets has resulted in most benchmark datasets containing annotations for only a limited number of structures, with all task-irrelevant regions labeled as background. This partial labeling creates significant challenges for model performance, particularly for multi-organ and tumor segmentation. The fragmented nature of these labels further limits model generalizability and scalability, often requiring separate models for different organs or tumors~\cite{yu2019crossbar,zhang2019light,zhu2019multi}, leading to redundant model development and increased operational complexity~\cite{xie2023learning}.

Designing a unified segmentation architecture capable of handling multi-organ and tumor segmentation has garnered considerable attention as a promising solution to these challenges. Multi-head segmentation models attempted to address these issues by using architectures with multiple decoders sharing a common encoder~\cite{chen2019med3d,fang2020multi,shi2021marginal}. However, training a separate decoder for each task prevented these models from effectively capturing semantic relationships between anatomical structures and pathological lesions. Dynamic models such as DoDNet and its variants introduced dynamic head architectures, where a controller generated convolutional parameters on the fly for adaptive segmentation~\cite{zhang2021dodnet,xie2023learning}. While effective in addressing the partial label problem, these models relied on one-hot vectors to specify the target class, ignoring inter-class semantic relationships that were critical for multi-organ and tumor segmentation. To overcome this, UniSeg replaced the one-hot vector with a learnable prompt and used bottleneck features to guide dynamic segmentation~\cite{ye2023uniseg}. However, limiting prompt learning to bottleneck features might constrain the model’s ability to capture multi-scale spatial information, which is essential for accurately distinguishing complex anatomical structures.

Unlike the aforementioned dynamic models, the CLIP-driven universal model utilized a pretrained CLIP text encoder~\cite{radford2021learning} to generate text embeddings, allowing it to capture semantic relationships between organs and tumors~\cite{liu2023clip}. This approach showed that incorporating anatomical context from CLIP text embeddings could enhance segmentation performance.  However, as the text embeddings interacted only with bottleneck features,  multiscale spatial information might not be fully captured. Additionally, dynamic convolution parameters were introduced only at the final stage of the decoder. This might limit the model’s generalizability. Moreover, existing dynamic models primarily relied on single-branch vision encoders, whose feature extraction capacity could be further enhanced to better integrate semantic and spatial information.


To address these challenges, we proposed a universal segmentation model that integrated self-supervised vision backbones with CLIP for medical image segmentation. Our model, the CLIP-DINO Prompt-Driven Segmentation Network (CDPDNet), belonged to the category of dynamic models but introduced three  innovations to better handle the partial labeling problem: 

\begin{itemize}

 \item In existing models~\cite{zhang2021dodnet,liu2023clip,ye2023uniseg}, one-hot labels, text embeddings, and learnable prompts interacted only with bottleneck features, neglecting fine-grained low-level features. This omission could lead to poor segmentation performance of small organs and tumors. In contrast, the proposed CDPDNet enabled text embeddings to interact with vision encoder features at multiple scales, enhancing the correspondence between text and image structure. By introducing text embeddings early in the network, the proposed model became “aware” of the task from the beginning of the inference process.
 \item The vision encoder utilized in the model combined both DINOv2 \cite{oquab2023dinov2}, a self-supervised Vision Transformer (ViT) pretrained on large-scale datasets, and a CNN to capture both fine-grained anatomical details and long-range dependencies, enabling robust and accurate segmentation across diverse anatomical structures.  
 \item In UniSeg~\cite{ye2023uniseg}, the learnable prompt lacked explicit text guidance, making it difficult to distinguish between tasks. Our model addressed this with a Text-based Task Prompt Generation (TTPG) module, which incorporated text guidance to generate more informative and task-specific prompts.
\end{itemize}

The proposed model was systematically evaluated on 11 segmentation benchmarks comprising 3,062 CT volumetric data, covering 25 organs, 6 tumors, and 1 cyst. To further assess generalizability, we also directly applied the trained model to two additional unseen datasets without fine-tuning. Experimental results show that the proposed segmentation network CDPDNet, which integrated self-supervised vision backbones and CLIP under the guidance of a text-based prompt, achieved substantial improvements in segmentation accuracy. 

\begin{figure*}[tp]
	\centering
	\includegraphics[width=18cm]{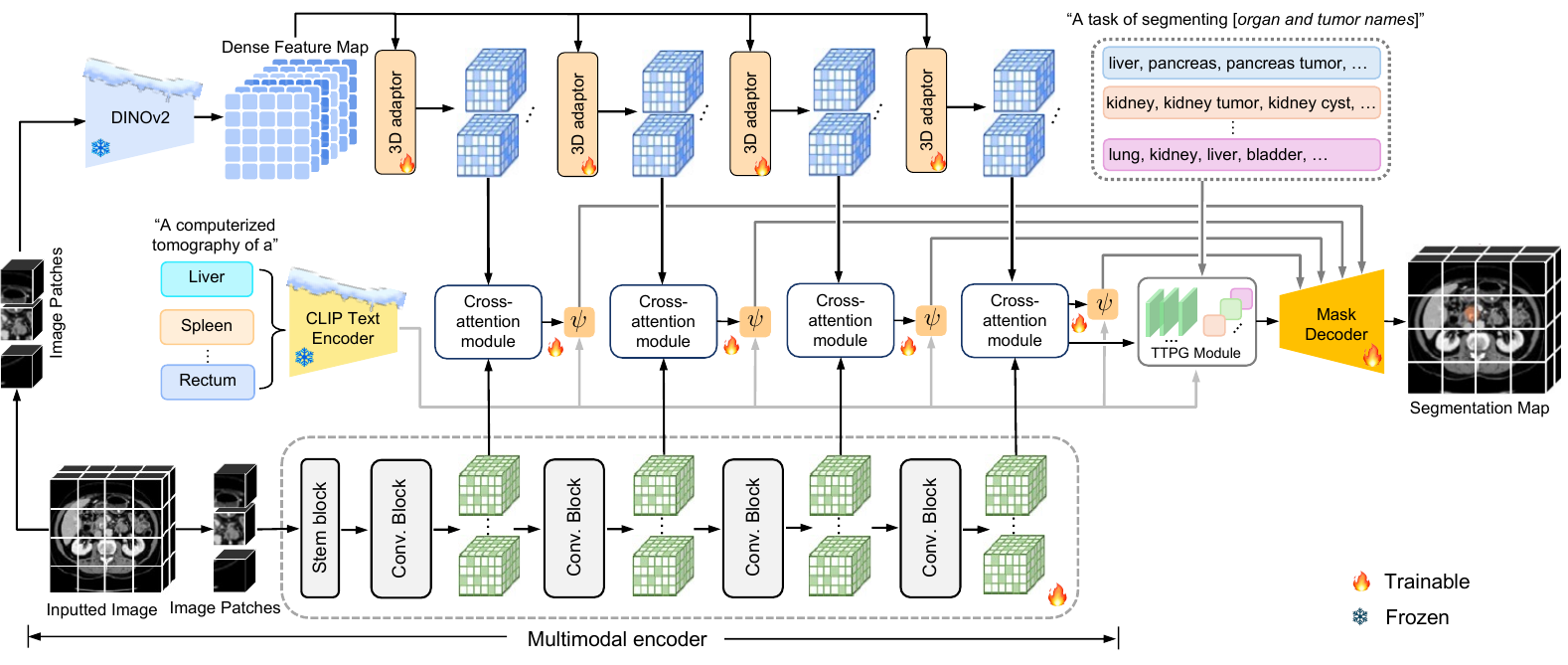}
	\caption{Overview of the proposed CLIP-DINO-Prompt Driven segmentation network (CDPDNet). It comprised three main components: a multimodal encoder integrating a DINOv2, a CLIP text encoder, and a CNN-based encoder (Sec.~\ref{sec:multiencoder}), a Text-based Task Prompt Generation (TTPG) module (Sec.~\ref{sec:ttpg}), and a mask decoder (Sec.~\ref{sec:decoder}). DINOv2 and CLIP text encoder extracted the dense visual and textual features. Vision features from DINOv2 and convolutional blocks were fused by leveraging cross-attention modules. Afterward, text features were aligned with the fused visual features using the alignment function $\psi$ (Sec.~\ref{clip}). Task-specific prompt was generated from the TTPG module and injected into the mask decoder to guide the final segmentation map prediction.}
	\label{fig1}
\end{figure*}

\section{Related Works}
\subsection{Partially Labeled Organ and Tumor Segmentation} Segmenting organs and tumors in abdominal imaging remains challenging due to the lack of large-scale, fully annotated datasets. Most publicly available datasets are limited to specific organs or tumors, resulting in partially labeled data. Early deep learning-based segmentation methods often addressed this by training separate models for each dataset, focusing on distinct structures. However, this approach compromises the overall efficiency and scalability.

To address this challenge, recent efforts have focused on developing a unified model across multiple datasets. Zhou et al.~\cite{zhou2019prior} introduced the Prior-aware Neural Network (PaNN), which first computed anatomical priors from a fully annotated dataset and later applied them to partially labeled data. Other methods have employed adaptive loss functions that can be directly applied to incomplete labels~\cite{fang2020multi,shi2021marginal,gonzalez2018multi}. From a network architecture perspective, Chen et al.~\cite{chen2019med3d} proposed a model featuring a shared encoder and task-specific decoders, trained across eight segmentation tasks. Huang et al.~\cite{huang2020multi} developed a co-training framework with cross-pseudo supervision, where two networks generated pseudo labels for each other. The prediction on unlabeled organs from one network was refined using the weight-averaged outputs of the other, facilitating unified multi-organ segmentation on few-organ datasets. In terms of dynamic models, Zhang et al.~\cite{zhang2021dodnet} reformulated the partially labeled segmentation task as a single-class problem and introduced DoDNet, a single-head network with dynamic weights for flexible organ and tumor segmentation. Building on this, Liu et al.~\cite{liu2023clip} proposed a CLIP-driven universal model that replaced DoDNet’s one-hot labels with CLIP text embeddings \cite{radford2021learning} to generate layer weights. Similarly, Ye et al. \cite{ye2023uniseg} introduced UniSeg, which employed a learnable prompt at the bottleneck instead of one-hot labels to guide segmentation. 

\subsection{Self-Supervised Vision Backbones}  
Recent advances in self-supervised learning have led to models with impressive capabilities in matching and localization tasks. Notably, the DINO series~\cite{caron2021emerging,oquab2023dinov2}, which utilized ViTs~\cite{dosovitskiy2020image} trained via a self-distillation loss function, have shown that patch-level representations trained through self-supervision could effectively capture semantic information. Moreover, the self-attention activations and foreground regions exhibited a strong relationship observed in the DINO-related models, sparking interest in their application to unsupervised segmentation tasks~\cite{simeoni2025unsupervised,wang2023tokencut,wang2022self,wang2023cut}. Regarding medical image segmentation,  a study conducted by Baharoona et al.~\cite{baharoon2023dino} demonstrated that DINOv2 outperformed other pre-trained models, particularly for radiology benchmarks.  Pérez-García et al.\cite{perez2025exploring} developed RAD-DINO by adapting the DINOv2 framework, showing that it surpassed several foundation models across multiple tasks, including classification, segmentation, and report generation. Building on these insights, our method integrated DINOv2 as a vision encoder to better extract multi-scale features and interact with CLIP text embeddings.

\section{Method}
\subsection{Problem Definition}
\label{ssec:problem}
Let $\mathcal{D} = \{ \mathcal{D}_1, \mathcal{D}_2, ..., \mathcal{D}_m \}$ denote a collection of $m$ partially labeled datasets, where each dataset  $\mathcal{D}_i=\{\bm{X}_i^j,\bm{Y}_i^j\}_{j=1}^{n_i}$ comprises $n_i$ pairs of image and corresponding ground-truth map. Each image $\bm{X}_i^j\in\mathbb{R}^{W\times{H}\times{D}}$ represents a volumetric scan with spatial dimensions $W\times{H}\times{D}$, while the corresponding ground-truth segmentation map $\bm{Y}_i^j\in\mathbb{N}^{W\times{H}\times{D}}$ contains discrete labels. Let $l_i=\{y|y\in{\bigcup\bm{Y}_{j=1}^{n_i}}\}$, then all the labels $L$ in $D$ can be represented as $L=\{y|y\in\bigcup_{i=1}^ml_i\}$. A partially labeled learning scenario arises when each individual dataset contains only a subset of the full label set, formally expressed as
$\forall i \in [1, m]$, $|l_i| < |L|$.

To solve this partially labeled problem, we treated the $m$ datasets as $m$ different segmentation tasks and the $i$-th task assigned a unique task identifier (ID) $i$. Then, this problem was reformulated as binary segmentation by leveraging CLIP-derived text embeddings $\bm{t}$, text-based task-specific prompt $\bm{t}_{task}$, and task ID $i$. Specifically, we transformed the original ground-truth segmentation map $\bm{Y}_i^j$ into a multi-channel tensor 
$\hat{\bm{Y}}_i^j\in\mathbb{N}^{W\times{H}\times{D}\times{|L|}}$ where each voxel $\hat{y}_i^j(h,w,d,c)$ was assigned as
\begin{equation}
	\hat{y}_i^j(h,w,d,c) = 
	\begin{cases}
		1 & \text{if label } c \text{ is present} \\
		0 & \text{if label } c \text{ is absent} \\
	\end{cases},
\end{equation}
where $(h,w,d,c)$ denoted the spatial location $(h,w,d)$ within the volume $\hat{\bm{Y}}_i^j$ with the corresponding label channel $c$. Given an input image $\bm{X}_i^j$, the CLIP text embeddings $\bm{t}$, the text-based task-specific prompt $\bm{t}_{task}$, and the task ID $i$, our objective was to train a unified segmentation network $\mathcal{F}$ that could predict all labels presented in $\bm{X}_i^j$. The model's prediction $\bm{P}_i^j$ could thus be formulated as
\begin{equation}
	\bm{P}_i^j=\mathcal{F}(\bm{X}_i^j,\bm{t}, \bm{t}_{task}, i;\bm{\theta}),
\end{equation}
where  $\bm{\theta}$ represented the learnable parameters of $\mathcal{F}$. Finally, the overall objective function for optimizing the model was defined as
\begin{equation}
	\min_{\bm{\theta}} \sum_{i=1}^m \sum_{j=1}^{n_i} \mathcal{L}(\mathcal{F}(\bm{X}_i^j,\bm{t},\bm{t}_{task}, i;{\bm{\theta}}), \hat{\bm{Y}}_i^j),
\end{equation}
where $\mathcal{L}$ denoted the segmentation loss function. 

\subsection{Architecture Overview}
Fig.~\ref{fig1} illustrates the architecture of the proposed CDPDNet, which consisted of a multimodal encoder, a Text-based Task Prompt Generation (TTPG) module, and a mask decoder. The network followed a U-shaped architecture that integrated a self-supervised vision backbone, DINOv2, with CLIP-based text embeddings, guided by a task-specific textual prompt. Specifically, the multimodal encoder comprised three branches: the top branch employed a pre-trained DINOv2 to extract dense visual features; the bottom branch was a vision encoder composed of convolutional blocks; and the middle branch included cross-attention modules to fuse visual features, along with alignment functions $\psi$ to align the text embeddings with the fused visual representations at different resolution levels. At the bottleneck stage, the TTPG module was designed to capture task-specific segmentation information. It enabled interaction between the task embeddings and the bottleneck features to generate a tailored prompt that guided the network toward task-aware segmentation. Finally, the fused features were passed through the mask decoder to produce the final segmentation map.
 
\subsection{Multimodal Encoder}
\label{sec:multiencoder}
\subsubsection{DINOv2-based Vision Encoder} 
\label{dino}
Existing dynamic models often employ convolutional layers as the building blocks. However, due to the limited receptive fields of convolutional layers,  capturing long-range dependencies is challenging. While recent approaches have introduced Swin Transformer blocks to mitigate this limitation, their fixed window sizes still constrain the ability to model global context. Moreover, although multi-head self-attention enhances representational power, it comes at the cost of substantial parameter overhead. To address these challenges, CDPDNet integrated a pretrained, self-supervised vision backbone, DINOv2, with a CNN-based vision encoder, enabling improved global context modeling without significantly increasing the number of parameters. In contrast to existing DINOv2-based segmentation methods~\cite{perez2025exploring}, which typically utilized embeddings from only the final layer, we extracted four groups of dense feature maps, $\bm{F}_D = \{\bm{F}_{D_k} \mid k = 1, \dots, 4\}$,  from the last four layers of DINOv2. These multi-scale features can provide richer and more fine-grained semantic representations.

Since DINOv2 was pre-trained on 3-channel images, the input 3D patches were first sliced into 2D axial slices, with each slice repeated three times to match the expected input format. However, this slicing-based approach failed to capture spatial dependencies along the axial direction. To overcome this limitation, we proposed a 3D adaptor that applied a depthwise convolution followed by a standard 3D convolution on the stacked dense feature maps, enabling effective aggregation of axial spatial information. Given a stacked dense feature volume at the $ k $-th level, denoted as $\bm{F}_{D_k} \in \mathbb{R}^{p_h \times p_w \times p_c} $, where $p_h$, $p_w$ and $p_c$ represented its spatial dimension, the output of the 3D adaptor, $ \bm{F}_{D_k}^{'} $ was computed as:
\begin{equation}
	\bm{F}_{D_k}^{'} = \operatorname{ReLU} \left( f_{\text{3D}} \big( f_{\text{depthwise}}(\bm{F}_{D_k}) \big) \right),
\end{equation}
where $ f_{\text{depthwise}} $ and $ f_{\text{3D}} $ represented the depthwise~\cite{chollet2017xception} and standard 3D convolution operations, respectively.

\begin{table}[t]
\centering
\renewcommand{\arraystretch}{1.2}
\caption{Architectural configuration of the CNN-based vision encoder.}
\resizebox{\columnwidth}{!}{%
    \begin{tabular}{c|c|c}
        \hline
        & \textbf{Feature Size} & \textbf{Kernel Configuration} \\
        \hline
        Input & $96 \times 96 \times 96 \times 1$ & -- \\
        \hline
        \multirow{2}{*}{Stem Block}
        & $96 \times 96 \times 96 \times 32$ & [$1\times3\times3$, $1\times1\times1$, 32 conv] \\
        & $96 \times 96 \times 96 \times 32$ & [$1\times3\times3$, $1\times1\times1$, 32 conv] \\
        \hline
        \multirow{2}{*}{Conv. Block \#1}
        & $96 \times 48 \times 48 \times 64$ & [$3\times3\times3$, $1\times2\times2$, 64 conv] \\
        & $96 \times 48 \times 48 \times 64$ & [$3\times3\times3$, $1\times1\times1$, 64 conv] \\
        \hline
        \multirow{2}{*}{Conv. Block \#2}
        & $48 \times 24 \times 24 \times 128$ & [$3\times3\times3$, $2\times2\times2$, 128 conv] \\
        & $48 \times 24 \times 24 \times 128$ & [$3\times3\times3$, $1\times1\times1$, 128 conv] \\
        \hline
        \multirow{2}{*}{Conv. Block \#3}
        & $24 \times 12 \times 12 \times 256$ & [$3\times3\times3$, $2\times2\times2$, 256 conv] \\
        & $24 \times 12 \times 12 \times 256$ & [$3\times3\times3$, $1\times1\times1$, 256 conv] \\
        \hline
        \multirow{2}{*}{Conv. Block \#4}
        & $12 \times 6 \times 6 \times 320$ & [$3\times3\times3$, $2\times2\times2$, 320 conv] \\
        & $12 \times 6 \times 6 \times 320$ & [$3\times3\times3$, $1\times1\times1$, 320 conv] \\
        \hline
    \end{tabular}%
}
\caption*{\footnotesize ``$1\times3\times3$, $1\times1\times1$, 32 conv" corresponds to a Conv-BN-LeakyReLU layer with a kernel size of $1\times3\times3$, stride of $1\times1\times1$ and 32 features.}
\label{tab1}
\end{table}

\subsubsection{CNN-based Vision Encoder} 
\label{cnn}
Although DINOv2 exhibited strong capabilities in capturing fine-grained local and global contextual information, its training on 2D natural images limited its effectiveness for 3D medical images. To mitigate this limitation, we incorporated an additional CNN-based vision encoder tailored for image feature extraction. Table~\ref{tab1} shows the parameter details of the CNN-based vision encoder. It consisted of a stem block followed by four convolutional blocks. The stem block included two stacked convolutions with a kernel size of $1 \times 3 \times 3$, mirroring the patch projection configuration used in DINOv2 for 2D images. The four convolutional blocks gradually reduced the spatial resolution and increased the channel size, similar to U-Net.


\subsubsection{Fusion of Vision Features} 

Given an input image $\bm{X}_i^j$, two sets of multi-level image features were extracted: $\bm{F}_D^{'}=\{\bm{F}_{D_k}^{'}|k=1,\dots,4\}$ from DINOv2 and $\bm{F}_S^{'}=\{\bm{F}_{S_k}^{'}|k=1,\dots,4\}$ from the CNN-based vision encoder. While $\bm{F}_D^{'}$ primarily captured global contextual semantics,  $\bm{F}_S^{'}$ focused on localized anatomical details. Due to this inherent difference in representation, direct feature fusion via concatenation was suboptimal.

To effectively bridge the gap between the two feature sets, we proposed a cross-attention mechanism that enabled adaptive alignment and integration of their complementary features. Specifically, the features from DINOv2, $\bm{F}_D^{'}$, were first upsampled to match the spatial dimensions of the CNN-based features, $\bm{F}_S^{'}$. Then, four cross-attention modules were applied at corresponding resolution levels to selectively highlight informative regions. At each level, given a query matrix $\bm{Q}_D$ derived from the DINOv2-based features and key-value matrices $\bm{K}_S$ and $\bm{V}_S$ from the CNN-based features, the multi-head cross-attention was calculated as

\begin{equation}
	\text{CA}(\bm{Q}_D, \bm{K}_S, \bm{V}_S) = \text{Softmax} \left(\frac{\bm{Q}_D \bm{K}_S^\top}{\sqrt{d}} \right) \bm{V}_S,
\end{equation}
where $d$ denoted the dimension of the query and key. 

\subsubsection{Text Embeddings and Vision Features Alignment} 
\label{clip}
CLIP-driven universal models have demonstrated that text embeddings based on organ and tumor names can capture anatomical relationships, thereby improving abdominal image segmentation~\cite{liu2023clip}. Building on this insight, we incorporated text embeddings into the proposed framework. While existing methods typically integrated text embeddings at the bottleneck layer, this might limit their alignment with features from small or fine-grained structures. To overcome this limitation, we aligned the text embeddings with multi-resolution visual features, enabling better interaction across anatomical scales.

As illustrated in the middle branch of Fig.~\ref{fig1}, we employed a pre-trained CLIP-based text encoder~\cite{radford2021learning}, the ViT-B/32 model, to generate text embeddings from the prompt: ``\texttt{A computerized tomography of a [ORGAN/TUMOR NAME].}" to capture the anatomical relationships. The generated text embeddings $\bm{t}\in\mathbb{R}^{\Omega_{\bm{t}}}$ were projected to the visual space $\mathbb{R}^{\Omega_{\bm{v}}}$ and aligned with the visual features $\bm{v}$ through an alignment function $\psi:\mathbb{R}^{\Omega_{\bm{t}}} \rightarrow \mathbb{R}^{\Omega_{\bm{v}}}$, defined as
\begin{equation} 
	\psi(\bm{v}, \bm{t}) = \left(\bm{W}_a \bm{t} + \bm{b}_a\right) \odot \bm{v} + \bm{W}_b \bm{t} + \bm{b}_b, 
    \label{eq6}
\end{equation} 
where $\bm{W}_a \in \mathbb{R}^{\Omega_{\bm{v}} \times \Omega_{\bm{t}}}$ and $\bm{W}_b \in \mathbb{R}^{\Omega_{\bm{v}} \times \Omega_{\bm{t}}}$ represented the learnable projection matrices, $\bm{b}_a$ and $\bm{b}_b$ denoted learnable bias vectors, and $\odot$ denoted the Hadamard product. 

\subsection{Text-based Task Prompt Generation}
\label{sec:ttpg}
In partially labeled organ and tumor segmentation, annotated regions vary across tasks. While some datasets include overlapping anatomical labels, others focus on distinct targets, posing challenges for unified model training. In Section~\ref{clip}, we leveraged text embeddings of organ and tumor names to capture semantic relationships across tasks. However, the correlations between each task and its annotated regions, as well as inter-task relationships, have not been explicitly modeled. To address this gap, we proposed the TTPG module, which generated task-aware prompts to help the model capture both intra-task and inter-task dependencies.

\begin{figure}[t]
	\centering
	\includegraphics[width=\linewidth]{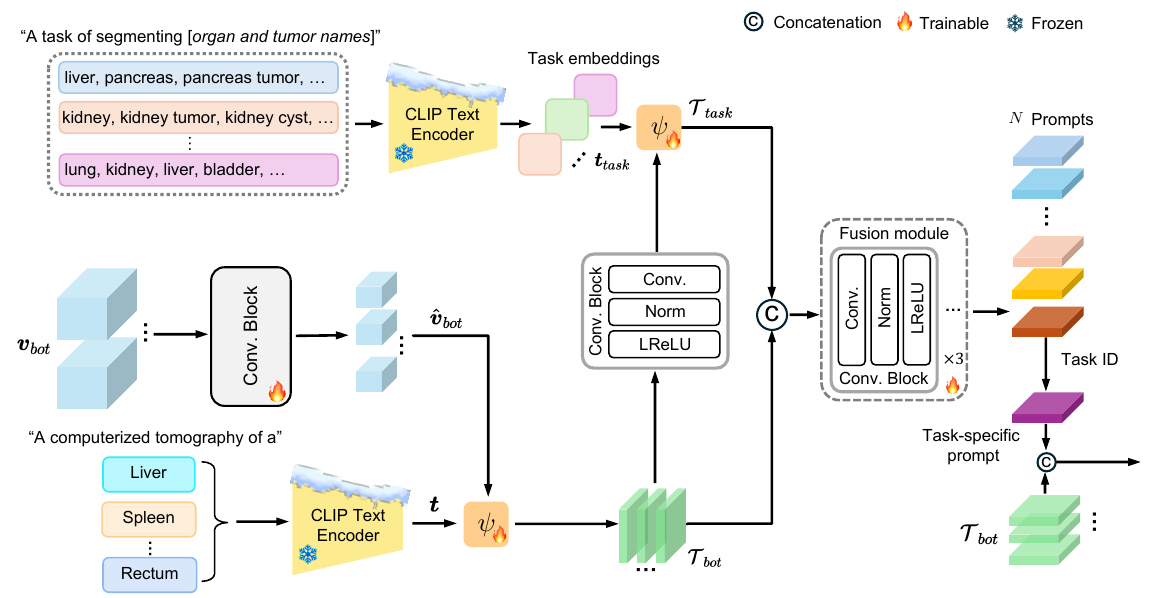}
	\caption{Architecture of the proposed Text-based Task Prompt Generation (TTPG) module.}
	\label{fig2}
\end{figure}

Fig.~\ref{fig2} shows the architecture of the TTPG module. For $N$ segmentation tasks, we constructed $N$ descriptive prompts in the form of: ``\texttt{A task of segmenting [ORGANS and TUMOR NAMES]}", where \texttt{[ORGANS and TUMOR NAMES]} was replaced with the actual target structures specific to each task. These prompts were encoded using the CLIP text encoder to obtain a set of task embeddings with dimension $N \times 512$. Let $\bm{v}_{bot}$ denote the bottleneck visual features and $\bm{t}$ the general text embeddings. The TTPG module first aligned $\bm{t}$ with a convolved version of $\bm{v}_{bot}$, $\hat{\bm{v}}_{bot}$, to form a joint vision-language representation, $\bm{\mathcal{T}}_{bot} = \psi(\hat{\bm{v}}_{bot}, \bm{t})$, using the alignment function $\psi$ defined in Eq.~(\ref{eq6}). This joint representation, $\bm{\mathcal{T}}_{bot}$, was further processed by a convolutional block and then fused with the task embedding $\bm{t}_{task}$ using the same function $\psi$, resulting in the task-specific representation $\bm{\mathcal{T}}_{task}$. Afterwards, $\bm{\mathcal{T}}_{task}$ was concatenated with $\bm{\mathcal{T}}_{bot}$ and fed into a fusion module to generate the task-aware spatial map. This map had the dimension of $B \times N \times h \times w \times d$, where $B$ denoted the batch size and $h \times w \times d$ represented the spatial dimension consistent with $\hat{\bm{v}}_{bot}$. The $i$-th channel of the spatial map was dynamically selected according to the task ID $i$ and treated as the task-specific prompt with a shape of $B \times 1 \times h \times w \times d$. After concatenation with $\bm{\mathcal{T}}_{bot}$, the combined representation was fed into the mask decoder to guide the segmentation process.

\begin{figure*}[t]
	\centering
	\begin{subfigure}{0.32\textwidth} 
            \hspace{0.05\textwidth}
		\includegraphics[width=0.9\linewidth]{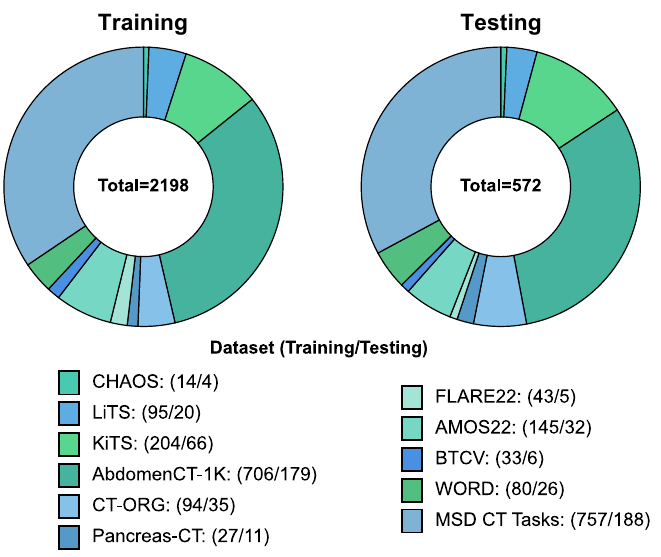} 
		\label{fig}
	\end{subfigure}
	\hfill 
	\begin{subfigure}{0.63\textwidth}
		\renewcommand{\arraystretch}{1.2} 
		\centering
		\resizebox{\textwidth}{!}{
			\begin{tabular}{lcl}
				\toprule
				\textbf{Datasets} & \textbf{Task ID} & \textbf{Annotated organs or tumors} \\  \midrule
				CHAOS        &1 & Liver \\
				LiTS         &2 & Liver, Liver Tumor* \\ 
				KiTS         &3 & Kidney, Kidney Tumor*, Kidney Cyst\\ 
				AbdomenCT-1K &4 & Spleen, Kidney, Liver, Pancreas \\ 
				CT-ORG       &5 & Lung, Kidney, Liver, Bladder \\ 
				Pancreas-CT  &6 & Spl, LKid, Gall, Eso, Liv, Sto, Pan, Duo \\ 
				FLARE22      &7 & Spl, RKid, Gall, Eso, Liv, Sto, Aor, Pos, Pan, RAG, LAG, CTr \\ 
				AMOS22       &8 & Spl, RKid, LKid, Gall, Eso, Liv, Sto, Aor, Pos, Pan, RAG, LAG, Duo, Bla, Pro \\ 
				BTCV         &9 & Spl, RKid, LKid, Gall, Eso, Liv, Sto, Aor, Pos, PSV, Pan, RAG, LAG, Duo \\ 
				WORD         &10 & Spl, RKid, LKid, Gall, Eso, Liv, Sto, Pan, RAG, Duo, Col, Int, Rec, Bla, LHF, RHF \\ 
				MSD CT Tasks &11 & Spl, Liv, Liv Tumor*, Pan, Pan Tumor*, Lun Tumor*, Col Tumor*, HV, HV Tumor* \\ 
				\bottomrule
			\end{tabular}
		}
		\label{tab}
	\end{subfigure}
	\caption{(a) Training and testing image composition. (b) Annotated 25 organs, 6 tumors, and a kidney cyst for 11 different segmentation tasks (datasets).}
	\label{fig3}
\end{figure*}

\subsection{Mask Decoder}
\label{sec:decoder}
The mask decoder comprised five sequential blocks, each consisting of one transposed convolution followed by two standard convolutions. Except in the final block, all transposed convolutions used a kernel size and stride of $2\times2\times2$ to progressively upsample the feature maps. To recover spatial resolution, the transposed convolution in the last block used a kernel size and stride of $1\times2\times2$. At each stage, the upsampled features were concatenated with the corresponding outputs from the alignment function $\psi$, and the combined features were processed by the subsequent convolutions. A final segmentation head with a $1 \times 1 \times 1$ kernel was used to generate the final segmentation map.

\subsection{Model Training} 
As depicted in Fig.~\ref{fig1} and Fig.~\ref{fig2}, the parameters of the DINOv2 and CLIP text encoders were frozen, while the parameters of the convolutional blocks, cross-attention modules, alignment function $\psi$, fusion blocks, and mask decoder were trainable. The trainable parameters were optimized using a combined loss function that incorporated both Dice loss and cross-entropy loss, expressed as:
\begin{equation} 
	\mathcal{L} = \alpha \mathcal{L}_{\text{Dice}} + \beta \mathcal{L}_{\text{CE}}, 
\end{equation}
where $\alpha$ and $\beta$ were the weighting factors that balanced the contributions of each loss component. In our study, $\alpha$ and $\beta$ were both set to 1. During training, we employed masked backpropagation, where the loss terms corresponding to absent classes in $\hat{\bm{Y}}_i^j$ were masked out, ensuring that only losses related to present classes were computed.

\section{Experiments}
\label{sec:experiment}
\subsection{Experimental Settings}

\subsubsection{Dataset} 
Our study utilized a dataset consisting of 3,062 CT volumetric images, divided into 2,198 for training, 292 for validation, and 572 for testing. This dataset included diverse CT volumetric images from 11 publicly available datasets: CHAOS~\cite{valindria2018multi}, LiTS~\cite{bilic2023liver}, KiTS~\cite{heller2020international}, AbdomenCT-1K~\cite{ma2021abdomenct}, CT-ORG~\cite{rister2020ct}, Pancreas-CT~\cite{roth2015deeporgan}, FLARE22~\cite{ma2023unleashing}, AMOS22~\cite{ji2022amos}, BTCV~\cite{igelsias2015miccai}, WORD~\cite{luo2022word}, and MSD CT tasks~\cite{antonelli2022medical}. The segmentation task covered a total of 32 regions of interest (ROIs), which included 25 abdominal structures, 6 types of tumors, and 1 cyst type. The 32 ROIs were as follows: spleen (Spl), right kidney (RKid), left kidney (LKid), gallbladder (Gall), esophagus (Eso), liver (Liv), stomach (Sto), aorta (Aor), postcava (Pos), portal vein and splenic vein (PSV), pancreas (Pan), right adrenal gland (RAG), left adrenal gland (LAG), duodenum (Duo), hepatic vessel (HV), right lung (RLun), left lung (LLun), colon (Col), intestine (Int), rectum (Rec), bladder (Bla), prostate (Pro), left head of femur (LHF), right head of femur (RHF), celiac trunk (CTr), kidney tumor (Kid Tumor), liver tumor (Liv Tumor), pancreatic tumor (Pan Tumor), hepatic vessel tumor (HV Tumor), lung tumor (Lun Tumor), colon tumor (Col Tumor), and kidney cyst (Kid Cyst). To test the generalizability of the model, another two datasets, including AbdomenCTCT \cite{hering2022learn2reg} and 3D-IRCADb~\cite{soler20103d}, were utilized. Details of each dataset used for training and testing were outlined in Fig.~\ref{fig3}.

\begin{table*}[t]
\centering
\renewcommand{\arraystretch}{1.1} 
\caption{The average DSC (\%) and HD for each of the 11 segmentation tasks, along with the overall averages of DSC and HD across all tasks, were obtained from 8 comparison methods. Values highlighted in red and blue indicate the methods that achieved the best and second-best performance, respectively.}
\resizebox{0.75\textwidth}{!}{%
		\begin{tabular}{ccccccccccccc}
			\toprule
			\multirow{2}{*}{Methods} & \multicolumn{2}{c}{CHAOS}   & \multicolumn{2}{c}{LiTS}   & \multicolumn{2}{c}{KiTS} & \multicolumn{2}{c}{AbCT-1K} & \multicolumn{2}{c}{CT-ORG}  & \multicolumn{2}{c}{PanCT} \\ \cline{2-13} 
			& DSC     & HD      & DSC     & HD     & DSC    & HD    & DSC     & HD      & DSC      & HD     & DSC    & HD          \\ \midrule
			UNet++              & 64.68   & 21.28   & 72.95   & 14.97  & 97.62  & 1.85  & 86.38   & 7.55    & 65.97    & 41.19  & 52.88  & 85.82       \\ 
			UNETR               & 69.37   & 16.55   & 71.48   & 15.41  & 97.57  & 1.93  & 79.06   & 38.18   & 62.72    & 43.03  & 64.24  & 17.08       \\ 
			Swin UNETR          & 72.15   & 17.44   & 72.51   & 15.63  & 97.59  & 1.85  & 84.96   & 6.19    & 71.75    & 35.72  & 69.06  & 15.73       \\ 
			TransUNet           & 72.30   & 16.99   & 71.37   & 16.09  & 97.82  & 1.64  & 87.25   & 6.62    & 68.11    & 33.10  & 73.48  & 9.57        \\ 
			CLIP-driven         & 75.64   & 17.58   & 71.48   & 16.66  & 97.74  & 1.72  & 88.36   & 5.43    & 72.40     & 29.04  & 72.04  & 11.33       \\ 
			CLIP-DoDNet            & 78.36   & \textcolor{red}{11.69}   & \textcolor{blue}{74.57}   & 14.57  & 97.71  & 1.68  & \textcolor{blue}{89.78}   & \textcolor{blue}{3.72}    & \textcolor{blue}{78.25}    & \textcolor{blue}{25.28}  & 75.26  & \textcolor{blue}{9.04}        \\ 
			PromptUniseg        & \textcolor{blue}{82.69}   & \textcolor{blue}{12.50}   & 74.29   & \textcolor{red}{13.85}  & \textcolor{blue}{97.85}  & \textcolor{red}{1.41}  & 88.86   & \textcolor{red}{3.54}    & 77.76    & 28.74  & \textcolor{blue}{76.55}  & 9.91        \\ 
			CDPDNet (Ours)                & \textcolor{red}{82.88}   & 19.03   & \textcolor{red}{74.92}   & \textcolor{blue}{14.02}  & \textcolor{red}{97.90}   & \textcolor{blue}{1.49}  & \textcolor{red}{90.33}   & 3.83    & \textcolor{red}{81.04}    & \textcolor{red}{15.67}  & \textcolor{red}{77.03}  & \textcolor{red}{8.38}        \\  \bottomrule
			\multirow{2}{*}{Methods} & \multicolumn{2}{c}{FLARE22} & \multicolumn{2}{c}{AMOS22} & \multicolumn{2}{c}{BTCV} & \multicolumn{2}{c}{WORD}    & \multicolumn{2}{c}{MSD-CTs} & \multicolumn{2}{c}{Avg.}  \\ \cline{2-13}   
			& DSC     & HD      & DSC     & HD     & DSC    & HD    & DSC     & HD      & DSC      & HD     & DSC    & HD          \\   \midrule
			UNet++              & 93.46   & 2.69    & 64.08   & 84.37  & 63.86  & 36.74 & 87.41   & \textcolor{red}{11.11}   & 73.26    & 19.66  & 74.78  & 29.75       \\ 
			UNETR               & 92.57   & 2.98    & 70.20   & 12.96  & 59.46  & 39.45 & 86.66   & 13.74   & 74.56    & 15.68  & 75.26  & 19.73       \\ 
			Swin UNETR          & 93.97   & 2.39    & 74.52   & 13.02  & 65.34  & 61.42 & 89.11   & 11.90   & 75.01    & 19.06  & 78.72  & 18.21       \\ 
			TransUNet           & 92.24   & 6.77    & 80.09   & 8.47   & 64.79  & 33.53 & \textcolor{blue}{89.30}   & \textcolor{blue}{11.31}   & 81.87    & 10.87  & 79.87  & 14.09       \\ 
			CLIP-driven         & 92.75   & 5.72    & 81.57   & 7.34   & 66.78  & 54.58 & 88.35   & 13.99   & 82.79    & 10.33  & 80.90  & 15.79       \\ 
			CLIP-DoDNet            & 94.16   & 2.50    & 82.28   & 7.53   & \textcolor{blue}{70.25}  & \textcolor{red}{28.20} & 88.51   & 15.97   & 82.71    & 10.08  & 82.89  & 11.84       \\ 
			PromptUniseg        & \textcolor{blue}{94.57}   & \textcolor{blue}{2.08}    & \textcolor{blue}{84.34}   & \textcolor{red}{5.79}   & 67.88  & 31.19 & \textcolor{red}{89.86}   & 12.90   & \textcolor{blue}{91.15}    & \textcolor{blue}{1.82}   & \textcolor{blue}{84.16}  & \textcolor{blue}{11.25}       \\ 
			CDPDNet (Ours)                & \textcolor{red}{94.75}   & \textcolor{red}{1.98}    & \textcolor{red}{84.35}   & \textcolor{blue}{5.91}   & \textcolor{red}{71.69}  & \textcolor{blue}{30.63} & 89.06   & 14.26   & \textcolor{red}{91.98}    & \textcolor{red}{1.60}   & \textcolor{red}{85.08}  & \textcolor{red}{10.62}       \\  \bottomrule
		\end{tabular}%
	}
	\label{tab2}
\end{table*}

\subsubsection{Implementation Details} 
The proposed method was implemented using PyTorch and MONAI 0.9. We employed the AdamW optimizer with an initial learning rate of $4 \times 10^{-4}$, a weight decay of $1 \times 10^{-5}$, and a momentum coefficient of 0.9. A linear warm-up cosine annealing learning rate scheduler was adopted, with a warm-up phase spanning the initial 50 epochs. The batch size was set to 1 per GPU, and the input patch size was fixed at $96 \times 96 \times 96$. Model training was conducted using Distributed Data-Parallel (DDP) across 8 NVIDIA A100 GPUs to ensure efficient multi-GPU scalability. For model selection, we identified the optimal checkpoint based on the highest segmentation accuracy achieved on the validation set. 

The data preprocessing pipeline involved clipping the intensity values to the range $[-175, 250]$, followed by normalization to $[0, 1]$. To ensure spatial consistency across scans, all volumes were resampled to achieve an isotropic voxel spacing of $1.5 \times 1.5 \times 1.5$ mm$^3$. Data augmentation strategies were employed to enhance the generalization capability of the training dataset. Specifically, a uniform sampling strategy was utilized to extract patches from each dataset with equal probability. Additionally, random 90-degree rotations and intensity shifts, applied with a probability ranging from 0.1 to 0.2, were incorporated to introduce variability.

\subsubsection{Evaluation Metrics}
To quantify the segmentation accuracy, we used a volume-based metric, Dice similarity coefficient (DSC), and a distance-based metric, Hausdorff distance (HD), as the evaluation metrics.

\subsection{Comparison with State-of-the-Art Methods}
\label{ssec:result}
To evaluate the performance of the proposed CDPDNet, we compared it with 6 state-of-the-art segmentation models: UNet++~\cite{zhou2018unet}, UNETR~\cite{chen2021transunet}, Swin UNETR~\cite{hatamizadeh2021swin}, TransUNet~\cite{chen2021transunet}, CLIP-driven Universal Model (CLIP-driven)~\cite{liu2023clip}, and Prompt-Uniseg~\cite{ye2023uniseg}. Additionally, since DoDNet \cite{zhang2021dodnet} can only be applied to segmentation tasks involving a single annotated organ/tumor or two annotated ROIs with an organ and its tumor, we compared our results with a modified version of DoDNet called CLIP-DoDNet. In CLIP-DoDNet, the one-hot vectors from the original DoDNet were replaced with CLIP-generated text embeddings from the CLIP-driven model. This adaptation enabled the model to segment all 32 ROIs considered in our experiments. All methods were evaluated under identical experimental settings for a fair comparison.

\begin{table*}[t]
	\centering
	\scriptsize
	\renewcommand{\arraystretch}{1.15} 
	\caption{The average DSC (\%) and HD for each of the 32 ROIs, along with the overall averages of DSC and HD across all ROIs, were obtained from 8 comparison methods. Values highlighted in red and blue indicate the methods that achieved the best and second-best performance, respectively.}
	\resizebox{\textwidth}{!}{ 
		\begin{tabular}{ccccccccccccccccccccccc}
			\toprule
			& \multicolumn{2}{c}{Spl}                                     & \multicolumn{2}{c}{RKid}                                   & \multicolumn{2}{c}{LKid}                                    & \multicolumn{2}{c}{Gall}                                    & \multicolumn{2}{c}{Eso}                                     & \multicolumn{2}{c}{Liv}                                     & \multicolumn{2}{c}{Sto}                                     & \multicolumn{2}{c}{Aor}                                     & \multicolumn{2}{c}{Pos}                                     & \multicolumn{2}{c}{PSV}                               & \multicolumn{2}{c}{Pan}                                     \\ \cline{2-23} 
			\multirow{-2}{*}{Methods} & DSC                          & HD                           & DSC                          & HD                          & DSC                          & HD                           & DSC                          & HD                           & DSC                          & HD                           & DSC                          & HD                           & DSC                          & HD                           & DSC                          & HD                           & DSC                          & HD                           & DSC                          & HD                           & DSC                          & HD                           \\  \midrule
			UNet++                    & 94.24                        & 2.18                         & 93.54                        & 3.46                        & 92.65                        & 4.38                         & 58.15                        & 43.27                        & 62.58                        & 23.46                        & 96.19                        & 3.16                         & 74.36                        & 17.67                        & 82.23                        & 14.95                        & 79.54                        & 14.14                        & 24.41                        & 59.27                        & 82.36                        & 5.13                         \\ 
			UNETR                     & 94.47                        & 2.09                         & 93.12                        & 3.45                        & 92.89                        & 3.45                         & 65.42                        & 12.63                        & 64.97                        & 11.20                        & 95.85                        & 3.56                         & 74.71                        & 17.96                        & 86.30                        & 6.39                         & 80.21                        & 6.06                         & 41.25                        & 29.35                        & 77.81                        & 6.88                         \\ 
			Swin UNETR                & 95.05                        & 1.82                         & 94.10                        & \textcolor{blue}{3.22} & 93.35                        & 3.65                         & 68.99                        & 16.77                        & 69.47                        & 10.05                        & 96.56                        & 2.50                         & 78.80                        & 14.89                        & 90.67                        & 3.93                         & 84.34                        & 7.75                         & 47.71                        & 40.31                        & 83.13                        & 4.57                         \\ 
			TransUNet                 & 95.04                        & 1.98                         & 85.81                        & 22.61                       & 93.27                        & 3.69                         & 73.46                        & 10.05                        & 67.50                        & 15.34                        & 96.65                        & 2.47                         & 79.57                        & 12.59                        & 91.40                        & 4.61                         & 85.90                        & 6.98                         & 64.15                        & 37.39                        & 84.58                        & 3.97                         \\ 
			CLIP-driven               & 95.26                        & 1.81                         & 87.14                        & 20.30                       & 93.88                        & \textcolor{blue}{3.04}  & 73.62                        & 13.39                        & 71.27                        & 11.43                        & 96.77                        & \textcolor{blue}{2.30}                         & 81.50                        & 11.91                        & 91.65                        & 4.18                         & 85.54                        & 7.67                         & 57.82                        & 41.79                        & 84.04                        & 3.99                         \\ 
			CLIP-DoDNet                  & 95.38                        & \textcolor{red}{1.64}  & 93.17                        & 5.62                        & 93.56                        & 3.55                         & \textcolor{blue}{77.55} & \textcolor{blue}{8.76}  & \textcolor{blue}{75.76} & \textcolor{blue}{7.74}  & 96.77                        & \textcolor{blue}{2.30}  & 80.15                        & 14.28                        & 92.76                        & 3.57                         & 87.28                        & 5.55                         & 62.37                        & 8.79                         & 85.24                        & 3.82                         \\ 
			PromptUniseg              & \textcolor{blue}{95.54} & \textcolor{blue}{1.75}                         & \textcolor{blue}{94.87} & 3.41                        & \textcolor{red}{94.54} & \textcolor{red}{2.69}  & 77.06                        & \textcolor{red}{5.79}  & \textcolor{red}{76.15} & \textcolor{red}{7.66}  & \textcolor{red}{96.82} & \textcolor{red}{2.25}  & \textcolor{blue}{86.31}                        & \textcolor{blue}{7.19}  & \textcolor{red}{92.84} & \textcolor{blue}{3.51}  & \textcolor{red}{88.00} & \textcolor{red}{3.75}  & \textcolor{blue}{64.95} & \textcolor{blue}{7.82}  & \textcolor{blue}{85.30} & \textcolor{blue}{3.78}  \\ 
			CDPDNet (Ours)                       & \textcolor{red}{95.88} & \textcolor{red}{1.64}  & \textcolor{red}{94.89} & \textcolor{red}{2.79} & \textcolor{blue}{94.23} & 3.24                         & \textcolor{red}{79.81} & 17.92                        & 74.38                        & 12.94                        & \textcolor{blue}{96.79} & 2.31                         & \textcolor{red}{87.32} & \textcolor{red}{6.94}  & \textcolor{blue}{92.80} & \textcolor{red}{3.47}  & \textcolor{blue}{87.65} & \textcolor{blue}{4.67}  & \textcolor{red}{66.28} & \textcolor{red}{7.50}  & \textcolor{red}{85.60} & \textcolor{red}{3.49}  \\ \toprule
			& \multicolumn{2}{c}{RAG}                                     & \multicolumn{2}{c}{LAG}                                    & \multicolumn{2}{c}{Due}                                     & \multicolumn{2}{c}{Col}                                     & \multicolumn{2}{c}{Int}                                     & \multicolumn{2}{c}{Rec}                                     & \multicolumn{2}{c}{Bla}                                     & \multicolumn{2}{c}{LHF}                                     & \multicolumn{2}{c}{RHF}                                     & \multicolumn{2}{c}{Pro}                                     & \multicolumn{2}{c}{HV}                                      \\ \cline{2-23} 
			\multirow{-2}{*}{Methods} & DSC                          & HD                           & DSC                          & HD                          & DSC                          & HD                           & DSC                          & HD                           & DSC                          & HD                           & DSC                          & HD                           & DSC                          & HD                           & DSC                          & HD                           & DSC                          & HD                           & DSC                          & HD                           & DSC                          & HD                           \\  \midrule
			UNet++                    & 10.79                        & 47.33                        & 9.76                         & 51.18                       & 61.95                        & 18.23                        & 48.50                        & 72.23                        & 54.93                        & 20.47                        & N/A                          & N/A                          & 51.81                        & 20.66                        & 83.57                        & 5.96                         & 3.28                         & 64.46                        & N/A                          & N/A                          & 56.70                        & \textcolor{red}{11.86} \\ 
			UNETR                     & 53.33                        & 13.02                        & 9.70                         & 54.86                       & 58.52                        & 14.70                        & 47.79                        & 72.08                        & 61.85                        & 14.88                        & 49.85                        & 7.26                         & 52.65                        & 21.61                        & 81.36                        & 7.18                         & 81.86                        & 5.70                         & 52.01                        & 18.83                        & 53.05                        & 27.08                        \\ 
			Swin UNETR                & 58.59                        & 14.54                        & 10.11                        & 54.97                       & 68.35                        & 12.91                        & 63.32                        & 54.38                        & 73.22                        & 9.56                         & 55.09                        & 19.69                        & 55.66                        & 20.84                        & 84.96                        & 5.78                         & 87.73                        & 4.67                         & 55.64                        & 30.47                        & 54.76                        & 21.43                        \\ 
			TransUNet                 & 66.07                        & 5.82                         & 51.02                        & 15.66                       & 67.65                        & 13.96                        & \textcolor{blue}{73.04} & \textcolor{red}{37.02} & 79.34                        & 8.71                         & 66.26                        & 4.57                         & 64.24                        & 13.79                        & 86.31                        & 4.24                         & 88.58                        & 7.72                         & 74.85                        & 8.38                         & \textcolor{blue}{55.88} & 15.30                        \\ 
			CLIP-driven               & 64.42                        & 6.56                         & \textcolor{blue}{65.26} & \textcolor{blue}{8.29} & 67.95                        & 15.52                        & 69.96                        & 46.40                        & 78.71                        & 9.45                         & 66.58                        & 4.44                         & 58.08                        & 19.25                        & 86.75                        & 4.11                         & 89.81                        & 4.29                         & 68.24                        & 13.24                        & 52.74                        & 18.36                        \\ 
			CLIP-DoDNet                  & 69.09                        & 3.58                         & 61.59                        & 13.02                       & 70.07                        & 14.33                        & 71.34                        & 43.97                        & \textcolor{red}{81.57} & \textcolor{blue}{7.88}  & 67.77                        & 4.04                         & 68.38                        & 14.07                        & 86.86                        & 3.95                         & \textcolor{red}{91.35} & \textcolor{red}{3.72}  & \textcolor{blue}{75.40} & \textcolor{blue}{6.16}  & 55.08                        & 12.09 \\  
			PromptUniseg              & \textcolor{blue}{70.16} & \textcolor{red}{3.32}  & 63.79                        & 12.82                       & \textcolor{blue}{71.09} & \textcolor{blue}{12.39} & 68.99                        & 52.93                        & \textcolor{blue}{81.20} & \textcolor{red}{7.75}  & \textcolor{blue}{68.87} & \textcolor{red}{3.92}  & \textcolor{red}{82.41} & \textcolor{blue}{6.71}  & \textcolor{blue}{87.33} & \textcolor{blue}{3.87}  & \textcolor{blue}{90.36} & \textcolor{blue}{4.26}  & \textcolor{red}{79.18} & \textcolor{red}{6.05}  & 55.40                        & 12.77                        \\  
			CDPDNet (Ours)                       & \textcolor{red}{70.33} & \textcolor{blue}{3.52}  & \textcolor{red}{68.46} & \textcolor{red}{4.99} & \textcolor{red}{71.85} & \textcolor{red}{11.84} & \textcolor{red}{74.48} & \textcolor{blue}{40.84} & 80.82                        & 11.56                        & \textcolor{red}{69.46} & \textcolor{blue}{3.93}  & \textcolor{blue}{81.73} & \textcolor{red}{6.64}  & \textcolor{red}{87.43} & \textcolor{red}{3.78}  & 88.95                        & 4.71                         & 74.52                        & 6.91                         & \textcolor{red}{56.54} & \textcolor{blue}{12.00} \\  \toprule
			& \multicolumn{2}{c}{Rlun}                                    & \multicolumn{2}{c}{Llun}                                   & \multicolumn{2}{c}{CTr}                                     & \multicolumn{2}{c}{Liv Tumor}                               & \multicolumn{2}{c}{Kid Tumor}                               & \multicolumn{2}{c}{Lun Tumor}                               & \multicolumn{2}{c}{Pan Tumor}                               & \multicolumn{2}{c}{HV Tumor}                                & \multicolumn{2}{c}{Col Tumor}                               & \multicolumn{2}{c}{Kid Cyst}                                & \multicolumn{2}{c}{Avg.}                                    \\ \cline{2-23} 
			\multirow{-2}{*}{Methods} & DSC                          & HD                           & DSC                          & HD                          & DSC                          & HD                           & DSC                          & HD                           & DSC                          & HD                           & DSC                          & HD                           & DSC                          & HD                           & DSC                          & HD                           & DSC                          & HD                           & DSC                          & HD                           & DSC                          & HD                           \\  \midrule
			UNet++                    & 76.95                        & 43.21                        & 95.38                        & 2.45                        & 86.08                        & 5.71                         & 76.37                        & 14.40                        & 48.33                        & 47.65                        & 51.01                        & 72.70                        & 35.36                        & 61.58                        & 58.81                        & 49.66                        & 22.19                        & 110.23                       & 30.37                        & 106.70                       & 56.32                        & N/A                          \\ 
			UNETR                     & 75.43                        & 47.23                        & 88.62                        & 13.82                       & 79.84                        & 4.38                         & 64.87                        & 42.60                        & 40.07                        & 59.26                        & 46.59                        & 60.56                        & 23.99                        & \textcolor{red}{37.76} & 50.20                        & 71.68                        & 25.10                        & 128.63                       & 25.81                        & 105.34                       & 62.17                        & 29.11                        \\ 
			Swin UNETR                & \textcolor{blue}{81.22} & \textcolor{blue}{40.24} & 89.79                        & 13.38                       & 89.33                        & 2.49                         & 74.12                        & 10.19                        & 60.52                        & 46.20                        & 52.90                        & 152.88                       & 33.97                        & 190.52                       & 61.48                        & 63.34                        & 34.52                        & 106.69                       & 38.31                        & 90.78                        & 68.30                        & 33.61                        \\ 
			TransUNet                 & \textcolor{red}{81.73} & \textcolor{red}{39.75} & 94.66                        & 3.81                        & 86.95                        & 3.41                         & 76.88                        & 11.33                        & 60.89                        & 51.11                        & 46.00                        & 53.20                        & 37.82                        & \textcolor{blue}{55.31} & 62.77                        & 66.51                        & 25.99                        & 93.34                        & 37.76                        & \textcolor{blue}{43.72} & 71.94                        & 21.20                        \\ 
			CLIP-driven               & 78.62                        & 43.04                        & 96.72                        & 2.32                        & \textcolor{blue}{89.98} & \textcolor{blue}{2.35}  & 79.96                        & 9.09                         & 56.91                        & 40.77                        & 52.65                        & 105.11                       & 39.27                        & 162.88                       & 63.71                        & 51.24                        & 35.71                        & 136.96                       & \textcolor{blue}{55.23}                        & 44.95                        & 72.99                        & 27.20                        \\ 
			CLIP-DoDNet                  & 75.07                        & 56.26                        & 92.24                        & 15.89                       & \textcolor{red}{91.03} & \textcolor{red}{2.11}  & \textcolor{blue}{82.13}                        & 6.09                         & \textcolor{blue}{73.32} & \textcolor{red}{22.16} & \textcolor{blue}{62.04} & \textcolor{blue}{50.67} & \textcolor{blue}{47.22} & 116.35                       & \textcolor{blue}{69.68} & \textcolor{red}{19.35} & \textcolor{blue}{36.97}                        & \textcolor{red}{42.12} & 50.58                        & 73.73                        & 75.59                        & \textcolor{blue}{18.66} \\ 
			PromptUniseg              & 72.81                        & 58.55                        & \textcolor{blue}{97.06}                        & \textcolor{red}{1.69} & 89.39                        & 2.99                         & 80.53                        & \textcolor{red}{5.12}  & 69.03                        & 50.94                        & 53.72                        & 107.65                       & 42.11                        & 63.60                        & 65.32                        & \textcolor{blue}{34.52} & 34.50                        & \textcolor{blue}{50.54} & 51.69                        & 59.00                        & \textcolor{blue}{75.85} & 19.09                        \\ 
			CDPDNet (Ours)                       & 67.49                        & 67.60                        & \textcolor{red}{97.24} & \textcolor{blue}{2.04} & 89.78                        & 2.48                         & \textcolor{red}{83.24} & \textcolor{blue}{5.88}  & \textcolor{red}{78.78} & \textcolor{blue}{36.05} & \textcolor{red}{66.30} & \textcolor{red}{40.62} & \textcolor{red}{48.45} & 68.82                        & \textcolor{red}{70.86} & 35.54                        & \textcolor{red}{40.29} & 106.71                       & \textcolor{red}{56.48} & \textcolor{red}{20.29} & \textcolor{red}{77.47} & \textcolor{red}{17.62} \\ 
			\bottomrule
		\end{tabular}%
	}
	\label{tab3}
\end{table*}

\begin{figure*}[t]
	\centering
	\includegraphics[width=18cm]{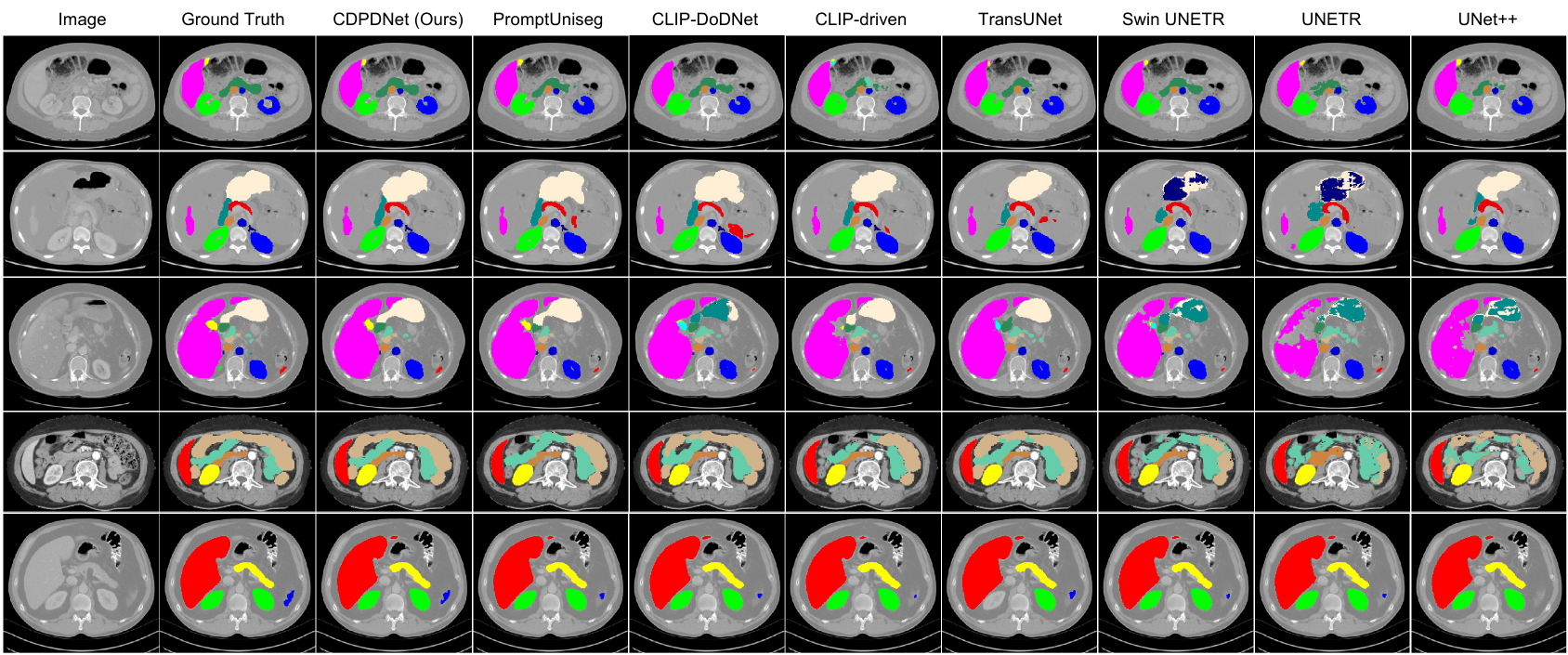}
	\caption{Visual comparison of segmentation methods on 5 representative organ segmentation samples from the testing dataset. The first column shows the image, and subsequent columns present results from ground truth and 8 comparison methods.}
	\label{fig4}
\end{figure*}

Experimental results were compared from two perspectives: across 11 different segmentation tasks and 32 distinct ROIs. Table~\ref{tab2} shows average DSCs and HDs of the 8 comparison methods for each of the 11 segmentation tasks, as well as the overall averages across all tasks. Except for the segmentation task WORD, CDPDNet achieved the best performance on the remaining 10 tasks in terms of DSC. Notably, on the CT-ORG and BTCV datasets, CDPDNet outperformed the second-best methods by more than 1.0\% in DSC. Regarding the HD values, both CDPDNet and PromptUniseg showed similar performance, as each achieved the lowest HD on 4 tasks. 
When considering the average DSC and HD values across the 11 tasks, CDPDNet outperformed the others, achieving the highest DSC (85.08\%) and the lowest HD (10.62). This demonstrated the superior performance of CDPDNet across a variety of tasks.

Table~\ref{tab3} presents the average DSC (\%) and HD values for each of the 32 ROIs and the overall averages across all ROIs. CDPDNet achieved the best overall performance with the highest average DSC of 77.47\% and the lowest average HD of 17.62. These results were notably better than the second-best methods, with PromptUniseg achieving a DSC of 75.85\% and CLIP-DoDNet yielding an HD of 18.66. This highlights the effectiveness of CDPDNet in segmenting both anatomical structures and pathological regions.

\begin{figure*}[!t]
	\centering
	\includegraphics[width=18cm]{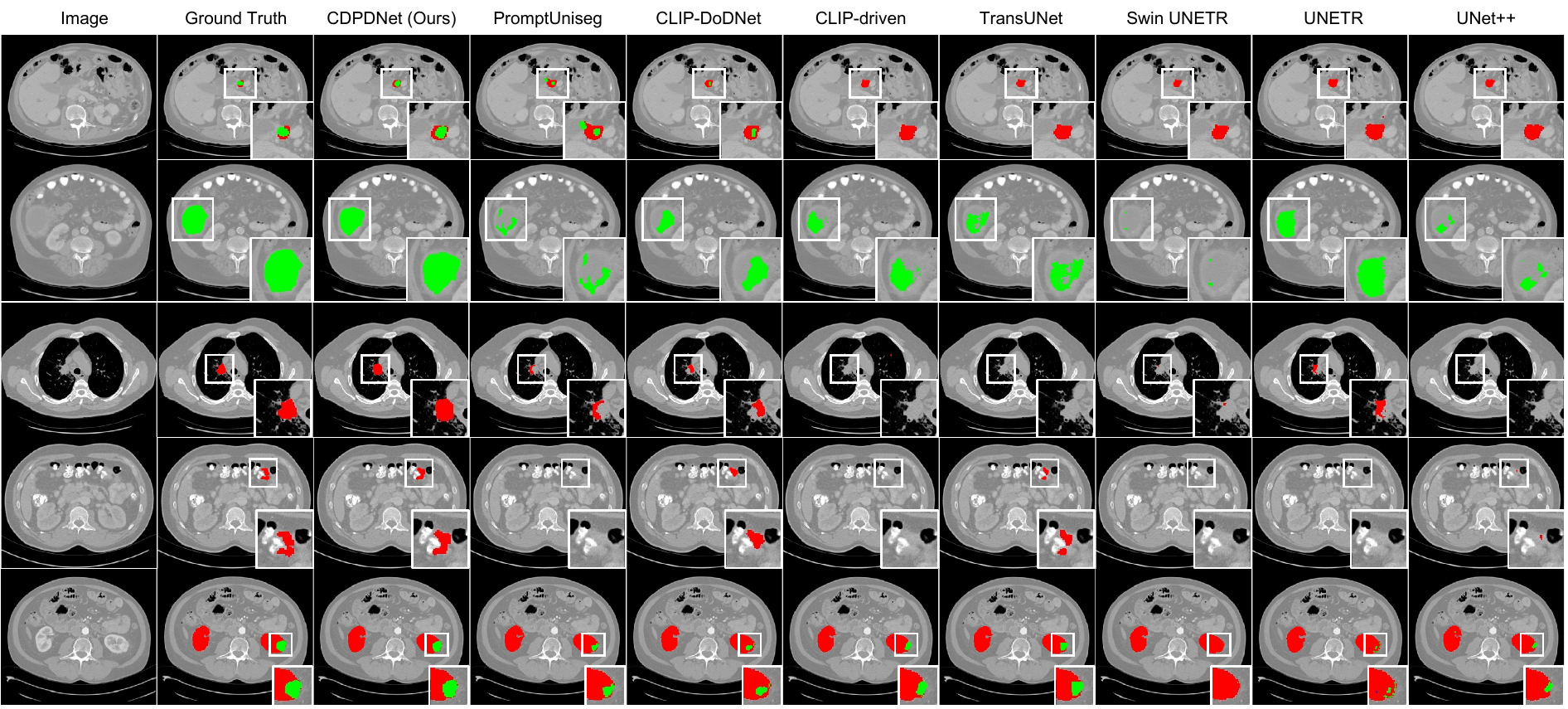}
	\caption{Visual comparison of segmentation methods on 5 representative tumor segmentation samples from the testing dataset. The first column shows the image, and subsequent columns present results from ground truth and 8 comparison methods.}
	\label{fig5}
\end{figure*}

Among the 25 anatomical structures, CDPDNet achieved the best and second-best DSC performance on 14 and 5 structures, respectively. Particularly, for the gallbladder (Gall), stomach (Sto), portal and splenic veins (PSV), left adrenal gland (LAG), and colon (Col), the DSC improvements over the second-best method were substantial ($>1.0\%$). For the six tumors and the kidney cyst, CDPDNet consistently achieved the highest DSC values. In particular, for the kidney tumor, lung tumor, and colon tumor, the margin of improvement was considerable ($\geq3.32\%$) compared to the second-best approach. These results underscore the capability of the proposed CDPDNet to deliver both accurate and balanced segmentation across a wide range of anatomical and pathological targets.

Fig.~\ref{fig4} shows anatomical structure segmentation maps of 5 representative samples obtained from 8 methods, along with the ground truth. It can be noted that CDPDNet achieved the closest segmentation results compared with the ground-truth maps. Although Prompt-Uniseg achieved segmentation results most similar to those of CDPDNet compared to the other 6 baseline methods, it generally failed to preserve fine boundary details of the segmented organs. For the remaining 6 baseline methods, issues such as mislabeling organs and failing to recognize certain structures were observed, particularly in the results from Swin UNETR, UNETR, and UNet++. Fig.~\ref{fig5} shows the segmentation results of the 8 comparison methods on 5 different tumor categories, including pancreatic tumor, hepatic vessel tumor, lung tumor, colon tumor, and kidney tumor (from top to bottom). As in organ segmentation, the proposed CDPDNet achieved the best performance in tumor segmentation, with results most consistent with the ground truth. In contrast, the other reference methods generally failed to identify the tumor regions accurately.

\subsection{Generalizability}
To evaluate the generalizability of the proposed CDPDNet, we conducted experiments on the AbdomenCTCT \cite{hering2022learn2reg} and 3D-IRCADb~\cite{soler20103d} datasets. Segmentation performance was assessed for 11 organs on the AbdomenCTCT dataset and for the liver tumor on the 3D-IRCADb dataset. Table~\ref{tab4} shows the average DSC (\%) for each of the 11 organs and the liver tumor, as well as the overall average DSC (\%) and average HD across all evaluated regions. The proposed CDPDNet achieved the best performance in terms of both DSC and HD on all organs and the liver tumor. Notably, it outperformed the second-best method by more than 10.0\% regarding DSC on the spleen, postcava, and pancreas. The overall average DSC and HD were significantly better than those of the second-best method (77.40\% versus 66.00\% and 13.54 versus 32.80, respectively). These results further demonstrate the superior performance of the proposed CDPDNet in segmenting both organs and tumors on unseen datasets.

\subsection{Ablation Studies}
\subsubsection{Effectiveness of the Main Components in the Proposed Framework}
To assess the contribution of the proposed components, we evaluated three key modules: the alignment module of CLIP-based organ-and-tumor-name embeddings and visual features (CLIP-EVA), the DINOv2-based dense feature extraction (DINOv2-DFE) module, and the TTPG module. We conducted ablation experiments on 8 model variants by incrementally integrating each individual component, all pairwise combinations, and all three components into a baseline model. The baseline model consisted solely of a CNN-based vision encoder and a mask decoder. The experimental datasets and settings were identical to those described in Sec.~\ref{ssec:result}. Table~\ref{tab5} summarizes the average DSC across all tasks and ROIs. Incorporating each of the three components consistently improved segmentation performance over the baseline. Among them, the TTPG module provided the largest performance gain, surpassing both the individual modules and their combination (CLIP-EVA and DINOv2-DFE), with an average DSC of 84.26\% versus 83.46\% across tasks, and 76.34\% versus 76.15\% across ROIs. When all three components were integrated, the proposed framework achieved the highest overall performance, demonstrating the  benefit of combining the proposed modules in CDPDNet.

\begin{table*}[t]
	\centering
	\scriptsize
	\caption{Average DSC (\%) and HD for 12 ROIs from AbdomenCTCT and 3D-IRCADb, along with overall averages across all ROIs, compared across 8 methods. Values highlighted in red and blue indicate the methods that achieved the best and second-best performance, respectively.}
	\resizebox{0.95\textwidth}{!}{%
		\begin{tabular}{ccccccccccccccc} \toprule
			Methods      & Spl                          & RKid                         & LKid                         & Gall                         & Eso                          & Liv                        & Sto                          & Aor                          & Pos                          & PSV                    & Pan                          & Liv Tumor                    & Avg. DSC                     & Avg.HD                       \\ \midrule
			UNet++       & 27.15                        & 87.61                        & 87.50                         & 49.51                        & 51.82                        & 88.53                        & 63.81                        & 64.68                        & 35.50                         & 25.25                        & 59.38                        & 61.48                        & 58.52                        & 66.32                        \\
			UNETR        & 42.51                        & 83.83                        & 85.28                        & 42.01                        & 38.41                        & 89.70                         & 67.55                        & 62.75                        & 24.80                         & 43.37                        & 29.36                        & 51.10                         & 55.06                        & 67.22                        \\
			Swin UNETR   & \textcolor{blue}{56.98} & 89.07                        & 87.09                        & \textcolor{blue}{63.67} & 48.48                        & 89.38                        & 73.52                        & 73.56                        & 34.58                        & 48.84                        & 49.29                        & 61.62                        & 64.67                        & 68.88                        \\
			TransUNet    & 33.05                        & 84.10                         & \textcolor{blue}{88.49} & 50.35                        & 58.65                        & 87.76                        & 82.72                        & 75.17                        & 31.28                        & 56.96                        & 57.17                        & 70.05                        & 64.65                        & 47.34                        \\
			CLIP-driven  & 16.37                        & 82.13                        & 87.98                        & 63.58                        & 49.31                        & \textcolor{blue}{90.80}  & 71.79                        & \textcolor{blue}{79.23} & \textcolor{blue}{42.16} & 57.17                        & 60.89                        & 66.53                        & 64.00                        & 43.18                        \\
			CLIP-DODNet     & 45.88                        & 90.90                         & 88.00                           & 61.39                        & 49.46                        & 88.49                        & 83.11                        & 56.59                        & 38.30                         & 55.78                        & 61.09                        & \textcolor{blue}{73.04} & \textcolor{blue}{66.00} & 45.34                        \\
			PromptUniseg & 16.29                        & \textcolor{blue}{90.76} & 88.45                        & 59.16                        & \textcolor{blue}{64.23} & 82.95                        & \textcolor{blue}{85.76} & 74.05                        & 34.42                        & \textcolor{blue}{61.71} & \textcolor{blue}{62.16} & 69.42                        & 65.78                        & \textcolor{blue}{32.80}                        \\
			CDPDNet (Ours)          & \textcolor{red}{80.88} & \textcolor{red}{91.26} & \textcolor{red}{89.22} & \textcolor{red}{67.88} & \textcolor{red}{67.53} & \textcolor{red}{93.73} & \textcolor{red}{89.83} & \textcolor{red}{82.27} & \textcolor{red}{52.87}  & \textcolor{red}{63.51} & \textcolor{red}{75.87} & \textcolor{red}{73.90}  & \textcolor{red}{77.40} & \textcolor{red}{13.54}
			\\ \bottomrule
		\end{tabular}%
	}
	\label{tab4}
\end{table*}

\begin{table}[t]
	\centering
	\scriptsize
	\caption{Ablation study of CLIP-EVA, DINOv2-DFE, and TTPG module. Performance is reported in DSC (\%) for both across Tasks and across ROIs settings.}
	\resizebox{\columnwidth}{!}{%
		\begin{tabular}{cccccc}
			\toprule
			Methods & CLIP-EVA & DINOv2-DFE & TTPG & Across Tasks & Across ROIs\\
			\midrule
			Baseline              & \ding{55} & \ding{55} & \ding{55} & 81.49 & 74.31 \\
			\multirow{7}{*}{CDPDNet \par (Ours)} 
			& \ding{51} & \ding{55} & \ding{55} & 82.64 & 75.03 \\
			& \ding{55} & \ding{51} & \ding{55} & 83.42 & 75.70 \\
			& \ding{55} & \ding{55} & \ding{51} & 84.26 & 76.34 \\
			& \ding{55} & \ding{51} & \ding{51} & 84.45 & 76.12 \\
			& \ding{51} & \ding{55} & \ding{51} & 84.53 & 76.36 \\
			& \ding{51} & \ding{51} & \ding{55} & 83.46 & 76.15 \\
			& \ding{51} & \ding{51} & \ding{51} & \textbf{85.08} & \textbf{77.47} \\
			\bottomrule
		\end{tabular}%
	}
	\label{tab5}
\end{table}

\subsubsection{Impact of Task ID Selection}
\label{subsec:taskid}
\begin{figure}[t]
	\centering
	\includegraphics[width=8.8cm]{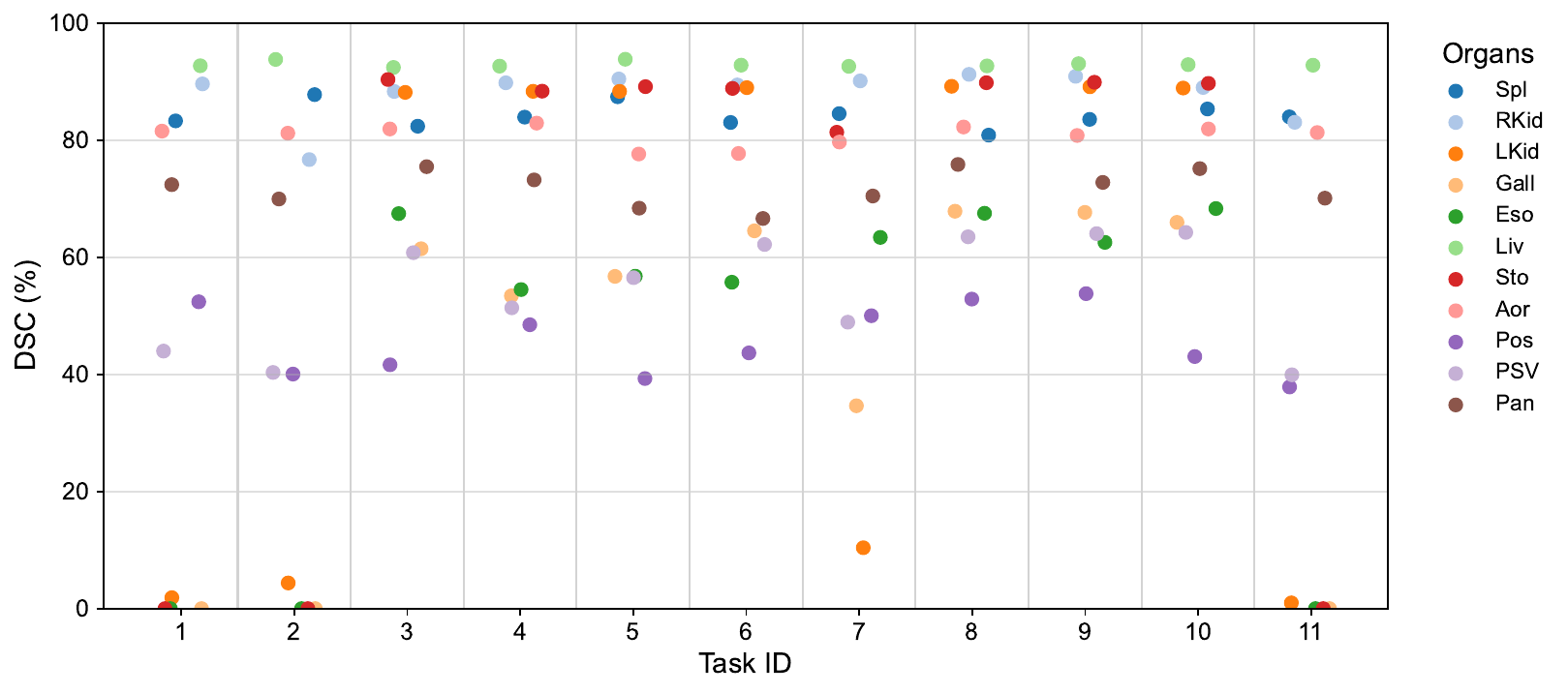}
	\caption{Average DSC distribution for each of the 11 organs on the AbdomenCTCT dataset using different task IDs. }
	\label{fig6}
\end{figure}

Task ID is a critical hyperparameter in the proposed method, particularly when segmenting datasets different from those used during training. To evaluate its effect, we conducted experiments on the AbdomenCTCT dataset using different task IDs. As shown in Fig.~\ref{fig6}, segmentation accuracy varied significantly with task IDs. IDs 1, 2, and 11, corresponding to the CHAOS, LiTS, and MSD CT tasks (see Fig.~\ref{fig3}), yielded poor performance due to limited overlap in annotated organs. In contrast, IDs 6, 8, 9, and 10 achieved markedly better results, as their training datasets covered most or all of the target organs. These results also indirectly demonstrate the effectiveness of the TTPG module in distinguishing between different tasks.

\section{Discussion}
\label{sec:discussion}

The proposed CDPDNet integrated text embeddings with visual features in a multi-scale manner, enabling more effective interaction between text embeddings and image features across different resolution levels. Experimental results show that the baseline model using CLIP-EVA outperformed the CLIP-driven method on the test set (see Table~\ref{tab3} and Table~\ref{tab5}), demonstrating the advantage of employing text embeddings that interacted with multi-scale visual features. Additionally, results show that the TTPG module provided explicit text guidance to generate task-specific prompts, which could achieve better segmentation outcomes than Prompt-Uniseg (see Table~\ref{tab3} and Table~\ref{tab5}). 

Another contribution of this work was the introduction of DINOv2 in conjunction with a CNN-based encoder for visual feature extraction. Although the DINOv2 parameters were kept frozen during training and inference, it had been pretrained on large-scale natural image datasets using a self-supervised learning strategy. This pretrained DINOv2 had the capability of capturing global contextual information and fine-grained anatomical details. To further adapt it for 3D medical image segmentation, we proposed a lightweight 3D adaptor, which could effectively bridge the domain gap without requiring fine-tuning of the full DINOv2 backbone. The effectiveness of this hybrid design was validated in the ablation studies (see Table~\ref{tab5}).

One limitation of this work is the need for manual selection of the task ID when applying the trained model to downstream segmentation tasks. One finding through the ablation study in Sec.~\ref{subsec:taskid} is that selecting the task ID associated with the dataset containing the target structures can lead to improved segmentation performance. In future work, we plan to explore an automatic task ID selection strategy by leveraging CLIP-based text embeddings, paving the way for a more user-interactive segmentation paradigm. Finally, in this work, we only tested the generalization capabilities of CDPDNet on two datasets, further studies of CDPDNet on larger, multi-site datasets are necessary to evaluate its robustness and real-world generalizability.

\section{Conclusion}
\label{sec:conclusion}
In this work, we proposed CDPDNet, a novel CLIP-DINO Prompt-Driven segmentation network designed for 3D medical image segmentation. CDPDNet effectively integrated self-supervised vision features with CLIP text embeddings, guided by task-specific prompts, to address the challenge of partial labeling in multi-organ and tumor segmentation. Key components of the proposed model included leveraging DINOv2 to cap long-range dependencies and enrich anatomical vision features, projection and alignment of CLIP-based organ and tumor name embeddings with vision representations, and the text-based task prompt generation for task-specific learning. Extensive experiments across diverse datasets demonstrate that CDPDNet achieved superior segmentation accuracy and robustness than other reference methods, highlighting its potential as a universal solution for medical image segmentation.

\bibliographystyle{IEEEtran}
\bibliography{reference.bib}

\end{document}